\documentclass[sigconf, nonacm]{acmart}
\AtBeginDocument{%
  }

\usepackage{times}
\usepackage{soul}
\usepackage{url}

\usepackage{caption}
\usepackage{amsmath,graphicx,hyperref}
\usepackage{amsthm}
\usepackage{booktabs}
\usepackage{algorithm}
\usepackage{algorithmic}
\usepackage{lscape}
\usepackage{booktabs}
\usepackage{multirow}
\usepackage{enumitem}
\usepackage{subcaption}
\usepackage{bbm}
\usepackage[mathbold]{mathfixs}

\usepackage[utf8]{inputenc}

\usepackage{multicol}
\newcommand{\best}[1]{\textbf{#1}}
\usepackage{makecell}

\renewcommand\footnotetextcopyrightpermission[1]{}
\begin{document}

\title{RECAST: Recent \& Context-Aware Sampling for Test-Time Adaptation in Streaming Biosignals}

\author{Yong-Yeon Jo}
\affiliation{%
  \institution{Medical AI Co., Ltd.}
  \city{Seoul}
  \country{Republic of Korea}
}

\author{Junho Song}
\affiliation{%
  \institution{Medical AI Co., Ltd.}
  \city{Seoul}
  \country{Republic of Korea}
}

\author{Joon-myoung Kwon}
\affiliation{%
  \institution{Medical AI Co., Ltd.}
  \city{Seoul}
  \country{Republic of Korea}
}

\renewcommand{\shortauthors}{Jo et al.}


\begin{abstract}

Streaming biosignals vary across subjects and drift over time,
so population-trained models lose accuracy during long-term monitoring.
Test-time adaptation (TTA) enables online personalization by updating the model on incoming samples.
But in a stream, a basic question is left open: \emph{which samples should drive each update?}
Using all buffered samples blurs the update with irrelevant segments.
Using only the latest segment makes the update noisy and unstable.
The most useful samples are recent, aligned with the current physiological state, and reliable enough to learn from.
We propose \textbf{RECAST} (REcent \& Context-Aware Sampling for TTA),
a lightweight sampling module for buffered TTA frameworks.
RECAST builds each adaptation batch from three signals:
temporal recency, contextual similarity, and predictive reliability.
It changes only which samples are used, leaving the model and the training objective unchanged.
On two blood-pressure datasets, RECAST improves estimation accuracy and trend tracking over baselines and ablations.
The per-patient gains are statistically significant on both datasets, with broad improvement on the regular benchmark and gains concentrated on the hardest patients in the emergency-department setting.
RECAST stays practical, adding only sub-second latency per segment on a single GPU and CPU core.

\end{abstract}

\ccsdesc[500]{Computing methodologies~Online learning settings}
\ccsdesc[300]{Information systems~Data streaming}
\ccsdesc[300]{Applied computing~Health informatics}
\ccsdesc[100]{Computing methodologies~Neural networks}

\keywords{Test-time adaptation, Streaming biosignals, Online personalization,
  Blood pressure estimation, Sample selection, Distribution shift,
  Continual learning}


\maketitle

\section{Introduction}
\label{sec:intro}

In real-world physiological monitoring, biosignals are collected continuously over long periods~\cite{lee2022vitaldb,gow2023mimic}, and unlabeled waveform segments arrive as a near-continuous stream, 
whereas reference measurements for calibration (e.g., cuff blood pressure) are available only intermittently~\cite{samsung_health_monitor,kansal2025mcmed} (see Figure~\ref{fig:scene-hybrid}).
In deployment, the goal is to continuously output reference-quality measurements (e.g., BP) from the waveform alone, providing predictions between these sparse references.
However, models trained on population data can become miscalibrated as patient characteristics and operating conditions shift over time. This calibration drift motivates continual recalibration and model updating in deployment~\cite{davis2020detectdrift}.

A central challenge is that, even within a single individual,
the incoming physiological stream changes over time in two characteristic ways
(see Figure~\ref{fig:data-anal})~\cite{wang2023pulsedb}.
First, \textbf{non-stationarity}:
gradual distribution drift makes older samples increasingly less representative of the current state,
motivating a recency bias in online updates.
Second, \textbf{abrupt transitions}:
physiological regimes can change rapidly, so recency alone is not sufficient;
the stream around a transition may contain mixed or unreliable segments.

One approach to adapting to such within-patient dynamics is test-time adaptation (TTA),
which updates a model online at deployment using test streams without full retraining~\cite{wang2020tent,niu2022eata,liu2021ttt++}.
Test-Time Calibration (TTC) extended TTA for personalized biosignals
by combining self-supervised adaptation with sporadic supervised calibration via a dual-buffer design~\cite{jo2025test}.
While TTC enables online personalization in a hybrid stream, sample selection remains under-specified.
Using all stored samples can blur updates with irrelevant segments,
while using only the latest segment leads to high-variance, noise-sensitive updates.
This raises a central question left open by TTC:
\emph{which} stored samples should drive each update.

To resolve this tension between blurred and high-variance updates,
we propose \textbf{RECAST} (\underline{RE}cent \& \underline{C}ontext-\underline{A}ware \underline{S}ampling for \underline{T}TA).
It is a lightweight sample selection module integrated into the TTC framework.
{RECAST} constructs each adaptation batch by combining three complementary criteria,
each addressing one of the stream dynamics above.
(i) \emph{Temporal recency}: it down-weights older samples via age-based exponential decay.
This directly counters non-stationarity,
where gradual drift makes old samples unrepresentative of the current state.
(ii) \emph{Contextual similarity}: it retrieves samples most similar to the current input
using cosine similarity in the encoder's latent space (formally introduced in Section~\ref{sec:recast}).
This handles abrupt transitions,
where recency alone would mix samples from the old and new regimes.
(iii) \emph{Predictive reliability}: it filters ambiguous candidates using uncertainty estimates.
This guards against the mixed segments that surround a transition,
keeping online updates stable.

The three criteria are not applied independently.
RECAST merges recency and contextual similarity into a single score per buffered sample,
ranking candidates that are both recent and aligned with the current input higher.
Predictive reliability then acts as a filter,
removing candidates whose predictions are too uncertain to update on safely.
Each adaptation batch is assembled from the surviving candidates,
keeping a fixed mix of labeled and unlabeled samples.
RECAST only changes which samples are used for each update.
The model and the TTC training objective stay the same.

We evaluated RECAST on two real-world datasets, PulseDB and MC-MED, that represent long-term monitoring with distinct temporal structures and label densities.
We compare RECAST with baselines and component ablations and report both accuracy- and tracking-oriented metrics.
We also provide analyses of per-patient improvement, within-patient variability, sampling behavior, hyperparameter sensitivity, and computational cost
to describe how the method behaves in streaming deployment.

Our results show that sample selection meaningfully affects the stability and effectiveness of streaming personalization, most clearly in regular streaming conditions.
On PulseDB, RECAST provides strong overall performance compared to baselines and ablations, which supports the benefit of combining recency, similarity, and reliability in regular streaming conditions.
On MC-MED, RECAST is again the most accurate method, though by a smaller margin, because most patients show little within-stay BP variation for adaptation to exploit.
Per-patient analysis shows that these improvements are statistically significant on both datasets.
They are broadest on PulseDB; on MC-MED the gains concentrate on the hardest and most variable patients.

\noindent{Our contributions are summarized as follows:}
\begin{itemize}
\item We identify sample selection as an under-specified but decisive component of buffered streaming TTA,
and frame it through two temporal issues: non-stationarity (drift) and abrupt transitions.
\item We propose RECAST, a \emph{lightweight plug-in sampling module} for buffered TTA frameworks
that combines temporal recency (age decay), contextual similarity (latent cosine), and predictive reliability (uncertainty filtering),
with no architectural changes to the host framework.
\item We show that RECAST yields statistically significant per-patient improvements on PulseDB and MC-MED,
and provide extensive analyses of sampling behavior, hyperparameter sensitivity, and computational cost,
where RECAST adds only sub-second per-segment latency.
\end{itemize}

\section{Related Work}
\subsection{Streaming Biosignal Setting}
\label{sec:data-flow}

\begin{figure}[t]
    \includegraphics[width=0.49\textwidth]{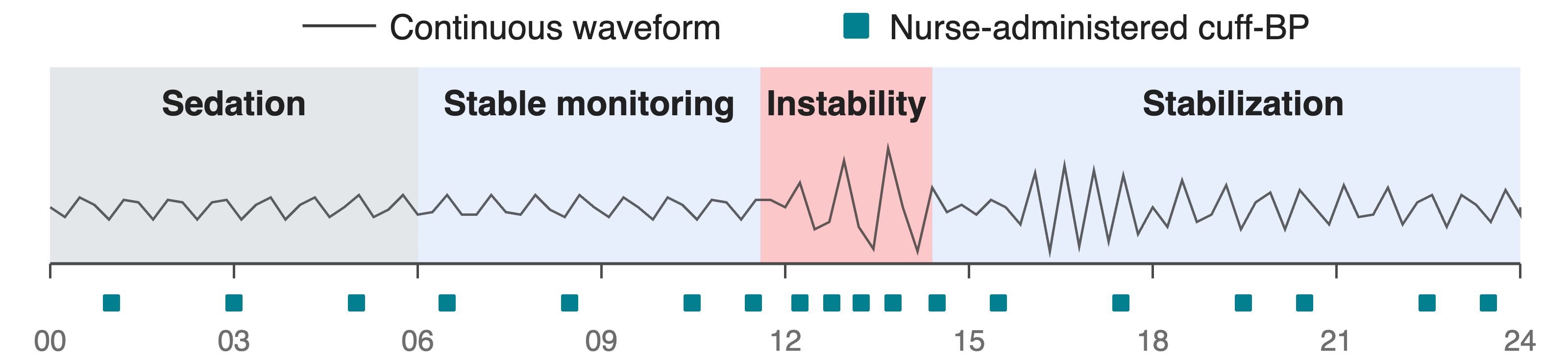} 
    \caption{
    Streaming biosignals at an ICU bedside over 24 hours. Waveforms (gray) arrive continuously while only sparse cuff-BP checks are available.
    }
    \label{fig:scene-hybrid}
\end{figure}

Biosignals can be ingested as continuous waveforms over long periods.
This occurs in clinical patient monitoring~\cite{kansal2025mcmed} and in everyday wearable devices~\cite{samsung_health_monitor}.
Alongside the waveform, reference measurements arrive only occasionally, to calibrate the signal,
such as cuff-based blood-pressure readings or finger-prick glucose tests~\cite{urden2013critical,galindo2020continuous}.
Figure~\ref{fig:scene-hybrid} illustrates a representative example of this setting: a single patient monitored at an ICU bedside over a 24-hour stay.
The gray trace is the continuously recorded biosignal, which dominates the stream.
The markers below it are the occasional cuff-based BP checks, the reference labels, taken only about once per hour.
The waveform also changes over the stay, with larger fluctuations during periods of physiological instability.
The example thus captures two defining traits of the setting at once: labels are sparse relative to the dense waveform, and the signal itself drifts over time within the same individual.
This creates a practical deployment challenge:
the model must track a continuously evolving stream,
while remaining anchored by occasional high-quality calibration labels.
Because the shift occurs primarily over time within the same individual, this setting differs from standard domain adaptation across populations.

\begin{figure*}[t]
    \centering
    \includegraphics[width=\linewidth]{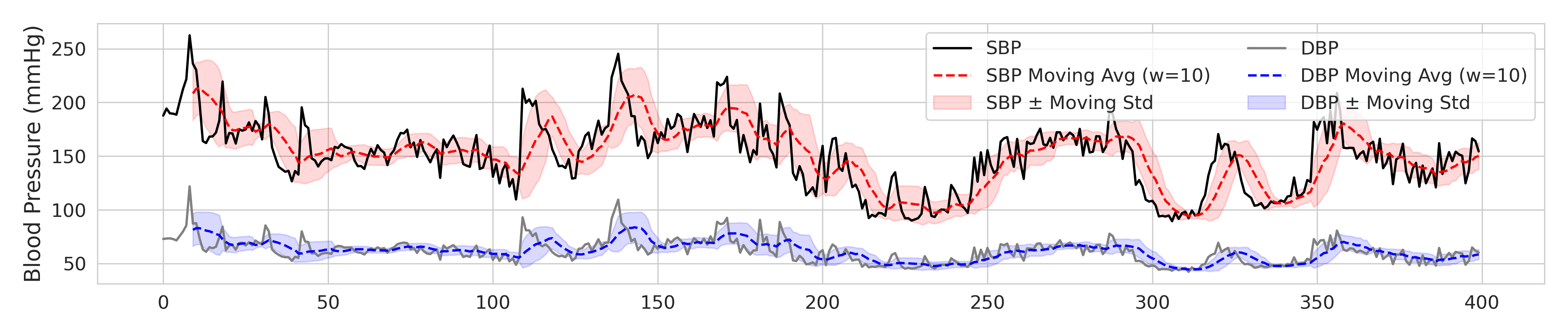}
    \caption{
     Blood pressure changes with moving statistics for a certain patient in PulseDB.
     The solid black and gray lines represent raw SBP and DBP, respectively.
     The dashed lines denote the moving average.
     The shaded regions represent the moving standard deviation.
    }
    \label{fig:data-anal}
\end{figure*}

\subsection{Test-Time Adaptation}

Test-time adaptation (TTA) updates a model online during inference to counter distribution shift, 
without revisiting training data~\cite{wang2020tent,liu2021ttt++}.
This makes it attractive for non-stationary, real-time deployment.

Existing methods differ in what they update and how.
Some adapt batch-normalization statistics or affine parameters by entropy minimization, as in TENT~\cite{wang2020tent}.
Others are selective about which samples to adapt on:
EATA~\cite{niu2022eata} and SAR~\cite{niu2023sar} filter unreliable, high-entropy or noisy samples.
A self-supervised family shares an auxiliary task between training and test time, including TTT~\cite{pmlr-v119-sun20b}, TTT++~\cite{liu2021ttt++}, and T-TIME~\cite{li2023t}.
TTT-MAE~\cite{gandelsman2022test} uses masked autoencoding~\cite{he2022mae} as the auxiliary task.

Most of these methods share two traits.
They target classification, where the core signal is the entropy of a softmax distribution.
And they adapt in a \emph{fully unsupervised manner, without using labels} that may arrive during deployment.

Test-Time Calibration (TTC)~\cite{jo2025test} extends test-time adaptation to personalized biosignals.
Unlike the classification-oriented methods above, it targets regression on streaming biosignals.
It combines self-supervised adaptation with periodic supervised calibration through a dual-buffer structure.

\section{Challenges in Adapting to Streaming Biosignals}
\label{sec:biosignal}

In real clinical monitoring, biosignals are measured under changing physiological conditions, such as rest, sleep, physical activity, and medical interventions~\cite{olufsen2005blood,parati2018blood}.
Because of these changes, the observation context shifts over time.
Figure~\ref{fig:data-anal} shows real recorded data from a single PulseDB patient.
We plot the patient's systolic and diastolic blood pressure (SBP, DBP) across their segments, together with the moving average and moving standard deviation.
The figure indicates that, even for a single individual, the signal distribution can change considerably over time.
Both a slow drift in the moving average and sharp, transient spikes are visible within the same recording.
This makes static models vulnerable in long-term monitoring.
In this paper, we focus on a practical challenge for streaming TTA: selecting and using buffered samples that match the current state.
We summarize biosignal dynamics into two issues and explain why common strategies are limited.

\paragraph{Issue 1 -- Non-stationarity: gradual distribution drift.}
Biosignals are non-stationary, meaning that basic statistics such as mean and variance can change slowly but continuously.
In Figure~\ref{fig:data-anal}, the moving average (dashed lines) shows gradual trends over time.
Such distributional drift is commonly addressed by emphasizing recent observations (e.g., sliding/fading windows or forgetting),
because older samples become outdated and less representative of the current physiology~\cite{gama2014survey,bayram2022concept}.
A simple and effective way is to apply temporal decay, where the contribution of a sample decreases, as the time elapsed since it was observed grows.
This helps the model stay aligned with the current distribution.

\paragraph{Issue 2 -- Abrupt transitions: rapid changes in state.}
In addition to gradual drift, abrupt transitions can occur due to sudden changes in activity or acute clinical events.
These transitions produce fast changes in signal characteristics within a short time window.
In Figure~\ref{fig:data-anal}, sharp SBP spikes and enlarged moving standard deviation (shaded regions) reflect such events.
In this case, using only the most recent samples can be risky.
The model should adapt using samples that are consistent with the new post-transition state.
Sometimes, a sample from earlier in the buffer can still be useful if it has a similar morphology to the current input.
For this reason, similarity-to-current is helpful to avoid misaligned updates during sudden regime changes.

These two issues pull in \emph{opposite directions}:
gradual drift rewards recency, while abrupt transitions can make the most recent samples misleading.
\emph{No single fixed rule resolves both.}
Yet existing TTA approaches rarely handle these dynamics explicitly.
In practice, buffer usage is often simplified to two extremes: updating only with the latest sample~\cite{liu2021ttt++} or with all buffered samples~\cite{jo2025test}.
The latest-only strategy can react quickly, but it is unstable and can overfit to noisy segments, potentially leading to forgetting.
Using all buffered samples is more stable during regular periods, but it can include outdated or context-mismatched samples, which harms adaptation.
This is why we weigh recency and similarity together, rather than relying on either extreme.
Even then, these factors act jointly, so the top-ranked sample is not always the safest to learn from.
As a safeguard, we add predictive reliability, which filters out candidates the model predicts with high uncertainty.

\section{Proposed Method: RECAST}
\label{sec:recast}
We now describe \textbf{RECAST}, a sampling module that addresses these challenges for streaming biosignal adaptation.
At each time step, RECAST builds an adaptation batch by considering three factors: temporal recency, contextual similarity, and predictive reliability.
The overall procedure is summarized in Algorithm~\ref{alg:recast}.

\begin{algorithm}[h]
\caption{RECAST: REcent \& Context-Aware Sampling for TTA}
\label{alg:recast}
\begin{algorithmic}[1]
\REQUIRE Current model $f_{t-1}$ with backbone $\Phi_{t-1}$, labeled queue $\mathcal{Q}_L$, unlabeled queue $\mathcal{Q}_U$, thresholds $(\tau_{\text{sim}},\tau_{\text{unc}})$, batch size $B$, label ratio $r$, decay rate $\lambda_t$, MC samples $S$
\FOR{each incoming test sample $x_t$ (optionally with $y_t$)}
    \STATE \textbf{Queue update:} $\mathcal{Q} \leftarrow \mathcal{Q} \cup \{x_t\}$ (and $y_t$ if available)
    \STATE Desired counts: $\bar{k}_L \leftarrow \lfloor rB \rfloor,\ \ \bar{k}_U \leftarrow B-\bar{k}_L$
    \STATE Initialize: $\mathcal{A}_L \leftarrow \emptyset,\ \mathcal{A}_U \leftarrow \emptyset$
    \FOR{each queue $\mathcal{Q} \in \{\mathcal{Q}_L,\mathcal{Q}_U\}$}
        \STATE $X_{\mathcal{Q}} \leftarrow \text{all samples in } \mathcal{Q}$
        \STATE $Z_{\mathcal{Q}} \leftarrow \Phi_{t-1}(X_{\mathcal{Q}})$ \COMMENT{Batched embedding of the entire queue}
        \STATE Identify current embedding $z_t$ and historical set $Z_{\text{hist}} \leftarrow Z_{\mathcal{Q}} \setminus \{z_t\}$
        \STATE Similarity vector: $s_{\mathcal{Q}} \leftarrow \mathrm{Sim}(z_t, Z_{\text{hist}})$
        \STATE Age vector: $a_{\mathcal{Q}} \leftarrow \mathrm{Age}(\mathcal{Q} \setminus \{x_t\})$
        \STATE Decayed similarity: $\tilde{s}_{\mathcal{Q}} \leftarrow s_{\mathcal{Q}} \odot \exp(-\lambda_t a_{\mathcal{Q}})$
        \STATE Uncertainty vector: $u_{\mathcal{Q}} \leftarrow \mathrm{Unc}(f_{t-1}, X_{\mathcal{Q}} \setminus \{x_t\}; S)$
        \STATE Age-penalized uncertainty: $\tilde{u}_{\mathcal{Q}} \leftarrow u_{\mathcal{Q}} \odot \exp(+\lambda_t a_{\mathcal{Q}})$
        \STATE Valid mask: $m_{\mathcal{Q}} \leftarrow (\tilde{s}_{\mathcal{Q}} \ge \tau_{\text{sim}})\ \wedge\ (\tilde{u}_{\mathcal{Q}} \le \tau_{\text{unc}})$
        \STATE $I_{\mathcal{Q}} \leftarrow \{ i \mid m_{\mathcal{Q}}[i]=\mathrm{True} \}$
        \STATE Sort $I_{\mathcal{Q}}$ by $\tilde{s}_{\mathcal{Q}}[i]$ descending
        \IF{$\mathcal{Q}=\mathcal{Q}_L$}
            \STATE $\mathcal{A}_L \leftarrow \{ x_i \mid i \in I_{\mathcal{Q}}[1{:}k] \}$ \COMMENT{$k \leftarrow \min(\bar{k}_L,\ |I_{\mathcal{Q}}|)$}
        \ELSE
            \STATE $\mathcal{A}_U \leftarrow \{ x_i \mid i \in I_{\mathcal{Q}}[1{:}k] \}$ \COMMENT{$k \leftarrow \min(\bar{k}_U,\ |I_{\mathcal{Q}}|)$}
        \ENDIF
    \ENDFOR
    \STATE $\mathcal{A} \leftarrow \mathcal{A}_L \cup \mathcal{A}_U$  
    \STATE Update model: ${f_{t}} \leftarrow \mathrm{Adapt}(f_{t-1};\mathcal{A})$
    \STATE Predict: $\hat{y}_t \leftarrow {f_{t}}(x_t)$
\ENDFOR
\end{algorithmic}
\end{algorithm}

We implement RECAST within the Test-Time Calibration (TTC) framework~\cite{jo2025test}.
TTC uses a multi-task learning structure with a \textit{shared backbone (encoder)} $\Phi: \mathcal{X} \to \mathbb{R}^d$ and two task-specific heads:
a \textit{reconstruction head} for self-supervised morphology restoration and a \textit{prediction head} for supervised blood pressure (BP) estimation.
We denote $z = \Phi(x) \in \mathbb{R}^d$ as the \emph{latent representation} of input $x$, and refer to the codomain of $\Phi$ as the \emph{latent space} in which RECAST's contextual similarity is computed.
During streaming inference, TTC maintains a \textit{queue-based dual buffer} $\mathcal{Q}=\{\mathcal{Q}_L,\mathcal{Q}_U\}$.
The labeled queue $\mathcal{Q}_L$ stores sparse ground-truth samples, while the unlabeled queue $\mathcal{Q}_U$ accumulates recent subject-specific waveform segments to reflect current dynamics.

\paragraph{Joint temporal-similarity scoring.}
RECAST scores each buffered sample to handle the two dynamics from Section~\ref{sec:biosignal}.
Under gradual drift, recent samples are more representative, so the score should favor recency.
After an abrupt transition, an older sample can still match the new state, so the score should also favor similarity.
Both features answer the same question: whether a buffered sample reflects the current state.
RECAST therefore fuses them into a single ranking score per buffered sample, rather than applying them as two separate rules.
We measure contextual similarity as the cosine similarity between latent embeddings from the current model $f_{t-1}$.
For each historical sample $x_i \in \mathcal{Q}\setminus\{x_t\}$, RECAST computes a \textit{joint temporal-similarity score} $\tilde{s}_i$:
\begin{equation}
    \tilde{s}_i = \mathrm{Sim}(z_t, z_i) \cdot \exp(-\lambda_t a_i),
    \qquad
    \mathrm{Sim}(z_t, z_i) = \frac{z_t^{\top} z_i}{\lVert z_t \rVert \, \lVert z_i \rVert},
\end{equation}
where $z_t=\Phi_{t-1}(x_t)$ and $z_i=\Phi_{t-1}(x_i)$ are embeddings from the backbone $\Phi_{t-1}$, $a_i$ is the age of the sample, and $\lambda_t$ is the decay rate.
The age term acts as a time-aware discount on similarity.
A sample ranks highly only when it is both similar and recent, and a low value on either signal pulls the score down.

\paragraph{Age-penalized uncertainty filtering.}
The joint score ranks buffered samples by relevance, but relevance alone does not guarantee a safe update.
As noted in Section~\ref{sec:biosignal}, a top-ranked sample can still be one the model predicts unreliably.
One concrete cause is a limitation of any compressed latent representation:
two samples can occupy nearby latent positions while arising from \emph{different} physiological states.
Such latent collisions are not detectable from cosine similarity alone,
and admitting these candidates would misguide adaptation.
We address this by filtering on predictive uncertainty, which serves as a complementary safeguard: latent collisions typically manifest as elevated epistemic uncertainty, because the encoder cannot map an ambiguous representation to a stable prediction.
We estimate epistemic uncertainty $u_i$ via Monte Carlo Dropout with $S$ stochastic forward passes of the current model $f_{t-1}$~\cite{gal2016dropout}.
With dropout activated at inference, each pass yields a different regression output $\hat{\mathbf{y}}_i^{(s)}$,
forming a set of stochastic predictions.
The dispersion of these predictions across the $S$ passes serves as an epistemic-uncertainty proxy.
We compute $u_i$ as the predictive variance over the $S$ predictions (averaged over output dimensions), and discard samples with large $u_i$.

To apply a temporal decay to uncertainty and enforce stricter criteria for older samples, we define an \textit{age-penalized uncertainty} $\tilde{u}_i$ using the decay rate $\lambda_t$:
\begin{equation}
    \tilde{u}_i = u_i \cdot \exp(+\lambda_t a_i).
\end{equation}
This penalty reduces the chance of adapting to uncertain and outdated patterns.
In streaming biosignals, it helps keep updates stable as the distribution drifts over time.

\paragraph{Adaptation batching.}
For each candidate sample $x_i$ in $\mathcal{Q}_L \cup \mathcal{Q}_U$, we mark it as valid for adaptation only if it passes both a similarity and an uncertainty threshold:
\begin{equation}
    m_i = (\tilde{s}_i \ge \tau_{\text{sim}}) \wedge (\tilde{u}_i \le \tau_{\text{unc}}).
\end{equation}
After filtering, RECAST forms an adaptation batch by maintaining a fixed labeled-to-total ratio $r$ (default $r=0.25$; see Appendix~B for the selection rationale).
We set the desired counts as $\bar{k}_L=\lfloor rB \rfloor$ and $\bar{k}_U=B-\bar{k}_L$.
Within each queue, we rank valid candidates by $\tilde{s}_i$ and select up to $\bar{k}_L$ samples from $\mathcal{Q}_L$ and up to $\bar{k}_U$ samples from $\mathcal{Q}_U$.
If fewer than the desired number of valid candidates are available, we use all valid samples, yielding $|\mathcal{A}|\le B$.
The final adaptation batch is $\mathcal{A}=\mathcal{A}_L \cup \mathcal{A}_U$.
We then update the model from $f_{t-1}$ to $f_t$ using the same reconstruction and prediction objectives as TTC, and output $\hat{y}_t=f_t(x_t)$.

Selecting on relevance alone admits matching but unreliable candidates, whose noisy predictions add update variance.
Selecting on reliability alone admits irrelevant candidates that drift the parameters toward stale regimes.
RECAST requires both, keeping only candidates that are relevant and reliable.

\section{Evaluation}
\label{sec:evaluation}

\begin{table*}[t]
\caption{Summary of datasets. \emph{Seg/visit} is the mean\,$\pm$\,std number of segments per visit. \emph{SBP} and \emph{DBP} are the population mean\,$\pm$\,std over all segments; \emph{within-pt std} is the median per-patient standard deviation of reference BP.}
\Description{Table summarizing the PulseDB and MC-MED datasets: number of visits and segments, segments per visit, population SBP and DBP mean and standard deviation, and within-patient BP standard deviation for the train and test splits. MC-MED shows larger across-patient spread but smaller within-patient BP variation than PulseDB.}
\label{tab:dataset-summary-2way}
\centering
\begin{tabular}{l|l|rr|c|cc|cc}
\toprule
\multirow{2}{*}{\textbf{Dataset}} & \multirow{2}{*}{\textbf{Split}} & \multirow{2}{*}{\textbf{\#Visits}} & \multirow{2}{*}{\textbf{\#Segments}} & \multirow{2}{*}{\textbf{Seg/visit}} & \multicolumn{2}{c|}{\textbf{BP (mmHg)}} & \multicolumn{2}{c}{\textbf{Within-pt std}} \\
\cmidrule(lr){6-7}\cmidrule(lr){8-9}
 & & & & & \textbf{SBP} & \textbf{DBP} & \textbf{SBP} & \textbf{DBP} \\
\midrule
\multirow{2}{*}{PulseDB} & Train & 2,506 & 902,160 & 360.0 & 118.60\,{\scriptsize$\pm$21.03} & 61.86\,{\scriptsize$\pm$12.65} & 12.4 & 6.8 \\
 & Test & 279 & 111,600 & 400.0 & 118.84\,{\scriptsize$\pm$20.59} & 62.00\,{\scriptsize$\pm$12.27} & 12.7 & 6.6 \\
\midrule
\multirow{2}{*}{MC-MED} & Train & 21,342 & 2,411,288 & 113.0\,{\scriptsize$\pm$84.4} & 131.00\,{\scriptsize$\pm$24.88} & 78.66\,{\scriptsize$\pm$17.13} & 8.0 & 6.4 \\
 & Test & 2,756 & 313,638 & 113.8\,{\scriptsize$\pm$86.5} & 131.87\,{\scriptsize$\pm$23.97} & 79.14\,{\scriptsize$\pm$16.90} & 7.6 & 6.1 \\
\bottomrule
\end{tabular}
\end{table*}

\begin{table*}[t]
\centering
\caption{Performance comparison of streaming TTA methods on PulseDB and MC-MED. \emph{MAE} and \emph{RMSE} are the mean absolute and root-mean-square errors for total, systolic (SBP), and diastolic (DBP) blood pressure; \emph{Std} is the error standard deviation for SBP and DBP (all lower is better); \emph{Corr.}\ is the Pearson correlation for SBP and DBP (higher is better). Bold indicates the best value within each dataset block.}
\Description{Table comparing streaming TTA methods on PulseDB and MC-MED. For each method we report total, SBP, and DBP MAE and RMSE, SBP and DBP error standard deviation, and SBP and DBP Pearson correlation. RECAST achieves the best MAE, RMSE, and correlation in both dataset blocks, with minor Std exceptions.}
\label{tab:main-results-expanded}

\begin{tabular}{l|l|ccc|cc|ccc|cc}
\toprule
\multirow{2}{*}{\textbf{Dataset}} & \multirow{2}{*}{\textbf{Method}} &
\multicolumn{3}{c|}{\textbf{MAE $\downarrow$}} &
\multicolumn{2}{c|}{\textbf{Corr. $\uparrow$}} &
\multicolumn{3}{c|}{\textbf{RMSE $\downarrow$}} &
\multicolumn{2}{c}{\textbf{Std $\downarrow$}} \\
& &
\textbf{Total} & \textbf{SBP} & \textbf{DBP} &
\textbf{SBP} & \textbf{DBP} &
\textbf{Total} & \textbf{SBP} & \textbf{DBP} &
\textbf{SBP} & \textbf{DBP} \\
\midrule

\multirow{4}{*}{PulseDB}
& No TTA
& 24.21 & 14.85 & 9.36 & 0.422 & 0.423
& 30.86 & 18.97 & 11.89 & 11.81 & 7.32 \\
& TTT-MAE
& 25.60 & 16.01 & 9.58 & 0.232 & 0.280
& 32.34 & 20.14 & 12.21 & 12.21 & 7.56 \\
& TTC
& 10.05 & 6.52 & 3.54 & 0.891 & 0.901
& 14.77 & 9.41 & 5.37 & \best{6.78} & 4.04 \\
& RECAST
& \best{9.51} & \best{6.20} & \best{3.30} & \best{0.896} & \best{0.909}
& \best{14.35} & \best{9.20} & \best{5.15} & 6.80 & \best{3.95} \\

\midrule

\multirow{4}{*}{\makecell[l]{MC-MED}}
& No TTA
& 29.52 & 17.53 & 11.99 & 0.460 & 0.342
& 38.78 & 22.73 & 16.05 & 14.47 & 10.66 \\
& TTT-MAE
& 31.58 & 19.16 & 12.42 & 0.238 & 0.232
& 41.32 & 24.88 & 16.44 & 15.87 & 10.77 \\
& TTC
& 20.04 & 11.63 & 8.41 & 0.778 & \best{0.700}
& 28.11 & 16.00 & 12.11 & 10.99 & \best{8.71} \\
& RECAST
& \best{19.44} & \best{11.22} & \best{8.22} & \best{0.791} & 0.699
& \best{27.72} & \best{15.68} & \best{12.04} & \best{10.95} & 8.80 \\

\bottomrule
\end{tabular}
\end{table*}

\subsection{Experimental Setup}
\subsubsection{Datasets.}
We evaluate on two publicly available multimodal clinical datasets with blood pressure (BP) labels: PulseDB~\cite{wang2023pulsedb} and MC-MED~\cite{kansal2025mcmed}.
Table~\ref{tab:dataset-summary-2way} summarizes their composition, BP statistics, and label density (BP Ratio).
The two datasets differ mainly in temporal structure and in how BP labels are supplied.

\textbf{PulseDB}~\cite{wang2023pulsedb} is a large-scale benchmark for cuff-less BP estimation built from MIMIC-III~\cite{johnson2016mimic} and VitalDB~\cite{lee2022vitaldb}.
It provides 10-second segments and includes only segments that already carry a reference SBP/DBP label, so every segment is labeled (100\% ratio).
We adopt the original patient-disjoint split and, to mimic realistic sparse supervision, stream with one BP label per ten segments (a \textbf{10\% ratio}) during adaptation.

\textbf{MC-MED}~\cite{kansal2025mcmed} is an emergency department dataset from PhysioNet with continuous physiologic waveforms and time-aligned clinical vitals from adult visits.
The dataset provides 60-second segments that arrive continuously, roughly once per minute, throughout a stay.
The number of segments per visit varies widely across patients (Table~\ref{tab:dataset-summary-2way}: $113.8 \pm 86.5$ in the test split, median 94), reflecting heterogeneous ED stay lengths.
PulseDB sessions, by contrast, are fixed-length.
BP values are not embedded in the stream but archived separately as intermittent measurements; aligning them to segments with simultaneous PPG and ECG covers only about 6\% of segments.
These references come from two sources: cuff (oscillometric) readings charted by clinicians, and values logged from the bedside monitor.
We evaluate only against the cuff readings (\textbf{about 2\% of segments}), because the cuff is the established clinical reference for blood pressure and matches the label type used in PulseDB; restricting to it keeps the evaluation target reliable and comparable across the two datasets.
Of the 2{,}756 test patients, 2{,}244 have at least one cuff reference and form the per-patient evaluation set.
We use the \emph{chronological} predefined split, which avoids patient overlap and places validation/test visits after training visits, matching our deployment-like setting where the model adapts on each patient's later visits.

The two datasets also differ in how BP varies (Table~\ref{tab:dataset-summary-2way}).
MC-MED spans a wider population BP range than PulseDB, reflecting a heterogeneous ED cohort.
Within a stay, however, each patient's BP is more stable: the median within-patient SBP std is $7.6$\,mmHg on MC-MED versus $12.7$ on PulseDB, whose surgical and ICU recordings fluctuate more during a session.
Since adaptation personalizes by tracking within-patient change, this leaves less for it to exploit on MC-MED.

\subsubsection{Model implementation.}
We build on the TTC framework~\cite{jo2025test}.
We use the same model for PulseDB and MC-MED, so comparisons stay consistent across the two supervision regimes.

\textbf{Input and preprocessing:}
Each streaming input is a fixed-length segment: 10\,s for PulseDB and 60\,s for MC-MED.
Segments are processed sequentially to simulate online monitoring.
All signals are resampled to 125\,Hz, bandpass filtered (1--60\,Hz), and normalized by sample-wise Z-score.

\textbf{Pre-training:}
The backbone is a 1D-CNN frontend followed by a Transformer encoder.
We pre-train the model using a self-supervised masked autoencoding objective with a masking ratio of 0.75.
We train for 50 epochs with Adam (learning rate $10^{-4}$), batch size 512, and mixed-precision training.
We apply gradient clipping for stable optimization.
The training loss is a weighted sum of prediction and reconstruction losses:
$\mathcal{L}=\lambda_{\mathrm{pred}}\mathcal{L}_{\mathrm{pred}}+\lambda_{\mathrm{recon}}\mathcal{L}_{\mathrm{recon}}$,
with $(\lambda_{\mathrm{pred}},\lambda_{\mathrm{recon}})=(1,10)$ so that reconstruction is prioritized.
For $\mathcal{L}_{\mathrm{pred}}$ we use Shrinkage Loss, and for $\mathcal{L}_{\mathrm{recon}}$ we use Smooth L1 Loss.
We select the best checkpoint by the lowest training loss.

\textbf{Test-time phase:}
We simulate streaming adaptation with a dual-buffer design ($\mathcal{Q}_L,\mathcal{Q}_U$), each with capacity 64.
For each incoming test segment, we form an adaptation batch of size $B=32$ with a labeled-to-total ratio $r=0.25$.
To personalize the representation while preserving prediction-head calibration, we update the shared backbone for 5 iterations per test input and keep the prediction head frozen.
Freezing the classifier anchors the latent-to-BP mapping to the population-trained prior, preventing subject-specific drift in the absolute prediction scale under sparse online supervision.
This encoder-only update regime also follows standard practice in test-time adaptation~\cite{wang2020tent,pmlr-v119-sun20b,gandelsman2022test,niu2022eata}, where the task head is typically held fixed to avoid catastrophic forgetting of the source-trained mapping.

We implement RECAST as the sample scoring module on top of TTC, using the best configuration (studied in Section~\ref{sec:ablation-hparams}) on each dataset.
The preprocessing steps above fully specify the input pipeline.
We estimate epistemic uncertainty via Monte Carlo dropout with $S{=}20$ stochastic forward passes.
All test-time updates use Adam with learning rate $10^{-4}$ and run on a single A100 GPU with one CPU core.

\subsubsection{Baselines and Ablations.}
\textsf{No TTA} is the source model without any test-time updates.
For each incoming segment, it runs a single forward pass with the population pre-trained weights and outputs the BP prediction directly.
No samples are buffered and no parameters are adapted, so it serves as the no-adaptation lower bound.
\textsf{TTT-MAE} is an unsupervised TTA baseline~\cite{gandelsman2022test}, adapted to our streaming setting.
At each step, the encoder is updated on a full batch of the most recent unlabeled samples from $\mathcal{Q}_U$.
It uses the same masked-autoencoding reconstruction objective as our backbone (mask ratio $0.75$), with the supervised calibration loss disabled.
\textsf{TTC} is the standard test-time calibration framework~\cite{jo2025test} that RECAST builds on.
It fills the adaptation batch with the most recent buffered samples, without any similarity- or reliability-based selection.

\subsection{Experimental Results}

\subsubsection{Performance Comparison.}
\label{sec:exp-sampling-strategy}

We first assess whether RECAST's selective adaptation improves overall streaming BP estimation, comparing it against the source model (No TTA), the unsupervised TTT-MAE baseline, and the standard TTC framework.
Table~\ref{tab:main-results-expanded} reports MAE and RMSE for total, SBP, and DBP, the error standard deviation (Std) for SBP and DBP, and Pearson correlation (Corr.) for SBP and DBP.
Total MAE is our primary accuracy metric, consistent with the per-patient analysis below.

On PulseDB, RECAST is the most accurate method, with the lowest MAE and RMSE on all three targets and the highest correlation for both.
It reduces Total MAE by $5.4\%$ over TTC ($9.51$ vs.\ $10.05$) and by $61\%$ over No TTA.
On MC-MED, RECAST again attains the lowest MAE and RMSE on all three targets and the best SBP correlation, reducing Total MAE by $3.0\%$ over TTC ($19.44$ vs.\ $20.04$) and by $34\%$ over No TTA.
Its margin over TTC is smaller here; we examine why in Section~\ref{sec:variability}.
The absolute errors are about twice as large on MC-MED as on PulseDB ($19.44$ vs.\ $9.51$ Total MAE), consistent with its more heterogeneous emergency-department cohort.

Beyond mean error, RECAST also tracks trends better.
Adaptation lifts SBP correlation from $0.42$ under No TTA to $0.90$ on PulseDB and from $0.46$ to $0.79$ on MC-MED, and RECAST holds the highest SBP correlation on both.
Its RMSE is lowest on every target as well, so the gains are not driven by a few easy segments.
The only two metrics where a baseline edges RECAST are SBP Std on PulseDB ($6.80$ vs.\ $6.78$) and DBP Std on MC-MED ($8.80$ vs.\ $8.71$), differences within rounding.

Two design choices underpin these gains, and both hold on PulseDB and MC-MED.
First, adaptation is essential: every TTC-based method cuts Total MAE sharply over No TTA, by about $59\%$ on PulseDB and $32\%$ on MC-MED.
Second, supervision is essential: the unsupervised TTT-MAE underperforms even No TTA on both datasets ($25.60$ vs.\ $24.21$ on PulseDB; $31.58$ vs.\ $29.52$ on MC-MED), and its correlation collapses (SBP $0.23$ on PulseDB).
Adapting the encoder by masked autoencoding alone, without TTC's intermittent supervised calibration, drifts away from the BP regression task.
RECAST adds scored sample selection on top of this supervised-hybrid framework, which yields its leading accuracy.

\subsubsection{Performance by Within-Patient Variability.}
\label{sec:variability}

This analysis explains why RECAST's margin is smaller on MC-MED: we test whether its benefit depends on how much a patient's BP varies within a stay.
Table~\ref{tab:variability} reports, per within-patient-variability quintile from Q1 (most stable) to Q5 (most variable), the fraction of patients improved and the median relative improvement on Total error, with the per-dataset counts.
We pool all evaluated patients from both datasets and sort them into these quintiles by within-patient SBP variability, the standard deviation of the reference SBP over a patient's segments.
The improvement is measured relative to TTC, $(\mathrm{MAE}_{\mathrm{TTC}}-\mathrm{MAE}_{\mathrm{RECAST}})/\mathrm{MAE}_{\mathrm{TTC}}$, which is comparable across the two datasets' different error scales.

\begin{table}[t]
\centering
\caption{Performance by within-patient SBP variability (Total error), pooled across PulseDB and MC-MED. Patients are sorted into quintiles Q1 (most stable) to Q5 (most variable) by within-patient SBP std. \emph{\% imp.}\ is the fraction of patients RECAST improves over TTC and \emph{gain} the median relative improvement.}
\Description{Per-quintile table relating within-patient SBP variability to RECAST's Total improvement. The most stable quintile, almost entirely MC-MED, improves about half of patients with about +0.9\% median gain; higher-variability quintiles reach 58 to 63\% improved with about +4\% median gain, where most PulseDB patients fall.}
\label{tab:variability}
\begin{tabular}{c|cc|cc}
\toprule
\multirow{2}{*}{\textbf{Quintile}} & \multicolumn{2}{c|}{\textbf{Total}} & \multicolumn{2}{c}{\textbf{\# patients}} \\
\cmidrule(lr){2-3}\cmidrule(lr){4-5}
 & \textbf{\% imp.} & \textbf{gain (\%)} & MC-MED & PulseDB \\
\midrule
Excluded & --- & --- & 852 & 0 \\
Q1 & 51.6 & $+0.9$ & 336 & 1 \\
Q2 & 57.5 & $+4.3$ & 310 & 22 \\
Q3 & 62.9 & $+4.4$ & 266 & 68 \\
Q4 & 62.9 & $+4.3$ & 244 & 90 \\
Q5 & 59.6 & $+3.8$ & 236 & 98 \\
\midrule
Total & & & 2{,}244 & 279 \\
\bottomrule
\end{tabular}
\end{table}

Overall, RECAST's gain rises with within-patient variability, so this characteristic, not the dataset itself, drives the benefit.
In the most stable quintile (SBP std $\approx3$\,mmHg), $51.6\%$ of patients improve with a $+0.9\%$ median gain.
From the second quintile on, the fraction improved rises to $58$--$63\%$ and the median gain to about $+4\%$.
The relationship is a threshold rather than a monotonic trend: a small amount of within-patient variation is enough to unlock the gain, which then plateaus and eases slightly in the most variable quintile ($59.6\%$ at Q5), where abrupt transitions leave fewer context-aligned samples for any buffered method to exploit.
On MC-MED, most patients sit in the low-variability quintiles, and it supplies almost all of the stable Q1 patients (336 vs.\ 1).
Its smaller aggregate margin therefore reflects this patient composition, not a failure of the method.
On PulseDB, patients fall almost entirely in the higher-variability quintiles, where the gain is largest, which is why its improvement is broad.
The per-target breakdown is in Table~\ref{tab:variability-target}, Appendix~\ref{app:variability}.

\subsubsection{Per-Patient Improvement Distribution.}

\begin{figure}[t]
\centering
\begin{subfigure}{0.49\linewidth}
\centering
\includegraphics[width=\linewidth]{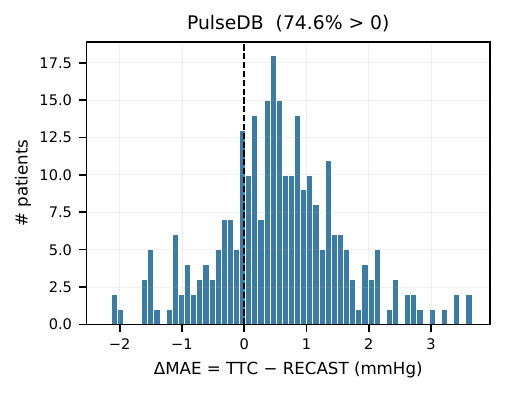}
\caption{PulseDB ($n=279$ sessions)}
\label{fig:per-patient-pulsedb}
\end{subfigure}
\hfill
\begin{subfigure}{0.49\linewidth}
\centering
\includegraphics[width=\linewidth]{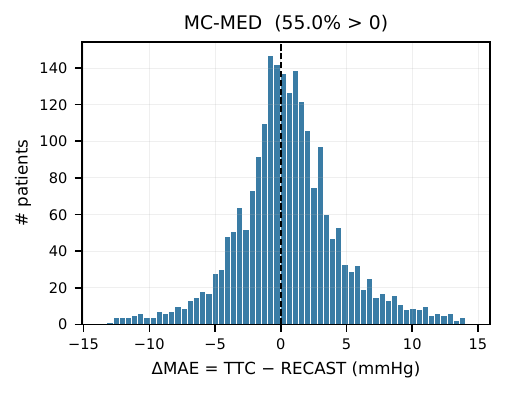}
\caption{MC-MED ($n=2{,}244$ patients)}
\label{fig:per-patient-mcmed}
\end{subfigure}
\caption{Per-patient Total $\Delta$MAE (MAE\textsubscript{TTC} $-$ MAE\textsubscript{RECAST}) distribution. Positive values on the $x$-axis indicate RECAST improvement.}
\Description{Histograms of per-patient Total $\Delta$MAE for PulseDB (left) and MC-MED (right). The horizontal axis is per-patient Total $\Delta$MAE (TTC minus RECAST), and bars to the right of the dashed zero line count patients that RECAST improves. On PulseDB most sessions lie to the right of zero; on MC-MED the distribution is more balanced, with a right-leaning tail of large gains.}
\label{fig:per-patient}
\end{figure}

Beyond the patient-averaged MAE in Table~\ref{tab:main-results-expanded}, we examine how the improvement is distributed across patients: how many patients RECAST helps, and which ones.
For each session (PulseDB) or patient (MC-MED) we measure the per-patient improvement $\Delta$MAE $=$ MAE\textsubscript{TTC} $-$ MAE\textsubscript{RECAST} over the standard TTC baseline.

Figure~\ref{fig:per-patient} plots the distribution of $\Delta$MAE for each dataset; bars to the right of the dashed line at zero are patients that RECAST improves.
RECAST improves the majority on both datasets, with broad and low-risk gains on PulseDB.
On PulseDB ($n{=}279$ test sessions), it reduces Total MAE for 74.6\% of sessions, with 29.4\% improving by more than 1\,mmHg and 8.2\% by more than 2\,mmHg; the 25th-percentile $\Delta$MAE stays near zero, so few sessions are made meaningfully worse.
On MC-MED ($n{=}2{,}244$ patients), 55.0\% improve, with 42.3\% improving by more than 1\,mmHg and 30.4\% by more than 2\,mmHg.
Paired Wilcoxon signed-rank tests confirm significant improvement on both datasets ($p\approx1.3\times10^{-16}$ on PulseDB; $p\approx1.4\times10^{-9}$ on MC-MED).

\begin{figure}[t]
\centering
\begin{subfigure}{0.49\linewidth}
\centering
\includegraphics[width=\linewidth]{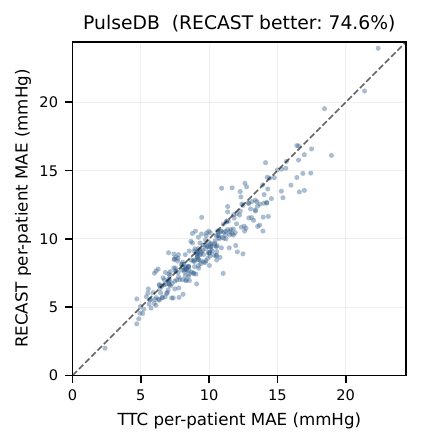}
\caption{PulseDB ($n=279$ sessions)}
\label{fig:pp-scatter-pulsedb}
\end{subfigure}
\hfill
\begin{subfigure}{0.49\linewidth}
\centering
\includegraphics[width=\linewidth]{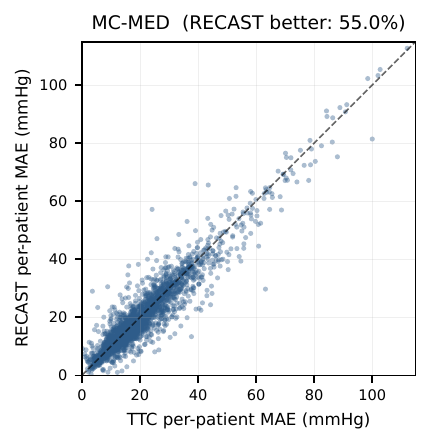}
\caption{MC-MED ($n=2{,}244$ patients)}
\label{fig:pp-scatter-mcmed}
\end{subfigure}
\caption{Per-patient Total MAE scatter: TTC on the $x$-axis vs.\ RECAST on the $y$-axis. Each point is one patient (or session). Points below the diagonal indicate RECAST improvement; points farther below indicate larger absolute gains.}
\Description{Scatter of per-patient Total MAE, TTC on the horizontal axis and RECAST on the vertical axis, for PulseDB (left) and MC-MED (right). Most points lie below the diagonal, more so on PulseDB, and the gap widens as the TTC error grows.}
\label{fig:pp-scatter}
\end{figure}

Figure~\ref{fig:pp-scatter} then relates each patient's RECAST outcome to its baseline difficulty: the per-patient TTC MAE on the $x$-axis against the RECAST MAE on the $y$-axis, with $y{=}x$ marking no improvement.
RECAST helps most where the population model is weakest.
Most points lie below the diagonal, more so on PulseDB, matching the per-dataset margins above.
They fall farther below as the baseline MAE grows: the harder a patient is for TTC, the larger the reduction RECAST tends to deliver.
On MC-MED this dependence is weaker.
The per-target (SBP and DBP) breakdown of both the distribution and the scatter follows the same pattern (Appendix~\ref{app:per-patient}).

\begin{figure*}[t]
  \centering
  \begin{subfigure}[t]{0.24\textwidth}
    \centering
    \includegraphics[width=\linewidth]{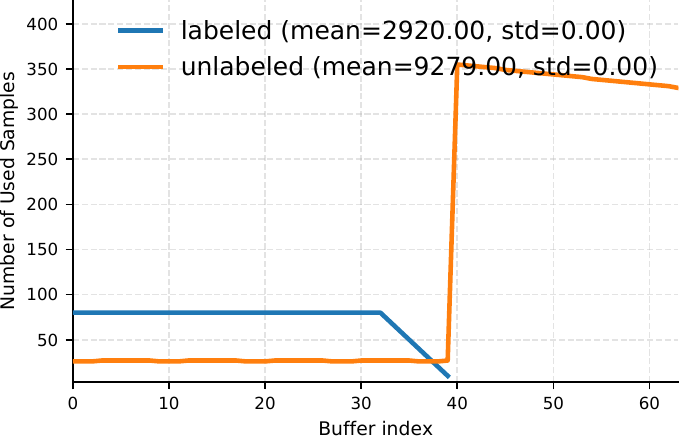}
    \caption{{TTC on PulseDB}}
    \label{fig:buf-usage-full-pulsedb}
  \end{subfigure}
  \begin{subfigure}[t]{0.24\textwidth}
    \centering
    \includegraphics[width=\linewidth]{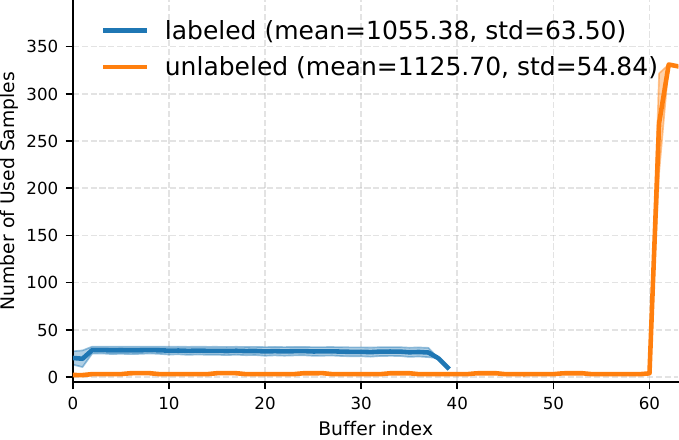}
    \caption{{RECAST on PulseDB}}
    \label{fig:buf-usage-recast-pulsedb}
  \end{subfigure} 
  \begin{subfigure}[t]{0.24\textwidth}
    \centering
    \includegraphics[width=\linewidth]{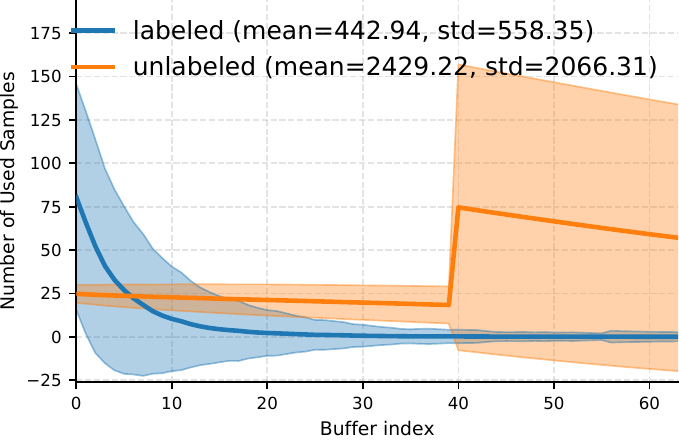}
    \caption{{TTC on MC-MED}}
    \label{fig:buf-usage-full-mcmed}
  \end{subfigure}
  \begin{subfigure}[t]{0.24\textwidth}
    \centering
    \includegraphics[width=\linewidth]{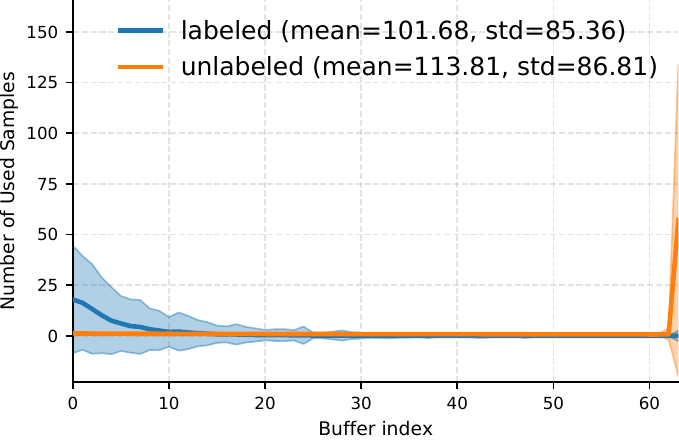}
    \caption{{RECAST on MC-MED}}
    \label{fig:buf-usage-recast-mcmed}
  \end{subfigure}
  \caption{Queue index-wise usage frequency of adaptation samples, aggregated over the entire test stream. Each panel plots how often each buffer position (FIFO queue index, $x$-axis) is selected into the adaptation batch ($y$-axis), shown separately for the labeled and unlabeled queues. The top row (a)--(b) is PulseDB and the bottom row (c)--(d) is MC-MED, each comparing TTC and RECAST. A flat profile indicates index-agnostic selection, whereas a dispersed (non-uniform) profile indicates preference for specific buffer regions.}
  \Description{For each method, a plot of how frequently each queue position is chosen for adaptation, shown separately for labeled and unlabeled queues; RECAST's distribution is more spread out than TTC's on both PulseDB and MC-MED.}
  \label{fig:buf-usage}
\end{figure*}

\subsubsection{Sampling Behavior Analysis.}

\label{sec:sampling-behavior}
To see how different TTA strategies form adaptation batches, we analyze the \emph{index-wise usage frequency} of buffered samples.
For each test-time update we record which queue positions enter the adaptation batch, and aggregate the counts over the whole test stream, separately for the labeled and unlabeled queues.
A near-uniform profile means the method is index-agnostic; a skewed profile means it prefers specific buffer regions.

Figure~\ref{fig:buf-usage} shows these frequencies aggregated over all visits.
Overall, RECAST samples in a \emph{context-dependent} way, unlike the fixed, recency-only selection of TTC, consistent with its accuracy gains in Table~\ref{tab:main-results-expanded}.

On PulseDB (Figures~\ref{fig:buf-usage}(a)--(b)), the stream is regular across visits.
Each session contributes about 400 segments and, at the 10\% label ratio, only about 40 labeled samples, so the labeled queue never fills its capacity of 64 and its usage spans roughly the first 40 indices.
TTC then produces an almost flat profile (Figure~\ref{fig:buf-usage}(a)), reflecting its deterministic, recency-based selection.
RECAST instead spreads its usage across indices (Figure~\ref{fig:buf-usage}(b)): because the similarity and uncertainty criteria depend on the current input, the selected set changes over time and shifts toward buffer regions that better match the current state.
On MC-MED (Figures~\ref{fig:buf-usage}(c)--(d)), the per-patient distribution of labeled and unlabeled samples is far more heterogeneous, and the state can shift abruptly with clinical interventions (\emph{Issue 2} in Section~\ref{sec:biosignal}).
Here even the TTC profile is not flat (Figure~\ref{fig:buf-usage}(c)), and RECAST remains selective (Figure~\ref{fig:buf-usage}(d)), placing weight on non-latest positions when the most recent samples are not representative of the current state.

\subsubsection{Hyperparameter Sensitivity.}

\begin{table}[t]
\centering
\caption{Ablation study on RECAST components and key hyperparameters on PulseDB. Best results in each block are boldfaced.}
\Description{Ablation table on PulseDB with four blocks: (A) temporal decay rate, (B) similarity threshold, (C) uncertainty threshold, and (D) component ablation. Each block reports total, SBP, and DBP MAE and SBP and DBP correlation. The chosen configuration attains the best values, and removing similarity, uncertainty, or using only the most recent sample degrades performance.}
\label{tab:ablation-pivot}
\begin{tabular}{c|ccc|cc}
\toprule
\textbf{Setting} & \multicolumn{3}{c|}{\textbf{MAE $\downarrow$}} & \multicolumn{2}{c}{\textbf{Correlation $\uparrow$}} \\
& \textbf{Total} & \textbf{SBP} & \textbf{DBP} & \textbf{SBP} & \textbf{DBP} \\
\midrule
\multicolumn{6}{l}{\textbf{(A) Temporal decay rate} $\boldsymbol{\lambda_t}$} \\
\midrule
\textbf{0.10} & \textbf{9.51} & \textbf{6.20} & \textbf{3.30} & \textbf{0.896} & \textbf{0.909} \\
0.05 & 10.00 & 6.51 & 3.49 & 0.889 & 0.902 \\
0.01 & 10.50 & 6.82 & 3.68 & 0.881 & 0.894 \\
\midrule
\multicolumn{6}{l}{\textbf{(B) Similarity threshold} $\boldsymbol{\tau_{\mathrm{sim}}}$} \\
\midrule
0.9 & 9.57 & 6.25 & 3.32 & 0.888 & 0.902 \\
\textbf{0.8} & \textbf{9.51} & \textbf{6.20} & \textbf{3.30} & \textbf{0.896} & \textbf{0.909} \\
0.7 & 9.83 & 6.41 & 3.42 & 0.891 & 0.904 \\
0.6 & 10.05 & 6.54 & 3.51 & 0.888 & 0.901 \\
0.5 & 10.34 & 6.73 & 3.60 & 0.883 & 0.897 \\
\midrule
\multicolumn{6}{l}{\textbf{(C) Uncertainty threshold} $\boldsymbol{\tau_{\mathrm{unc}}}$} \\
\midrule
\textbf{0.9} & \textbf{9.51} & \textbf{6.20} & \textbf{3.30} & \textbf{0.896} & \textbf{0.909} \\
0.8 & 9.55 & 6.22 & 3.32 & 0.895 & 0.907 \\
0.7 & 9.62 & 6.27 & 3.34 & 0.893 & 0.906 \\
0.6 & 9.66 & 6.30 & 3.36 & 0.893 & 0.905 \\
0.5 & 9.75 & 6.36 & 3.39 & 0.890 & 0.904 \\
\midrule
\multicolumn{6}{l}{\textbf{(D) Component ablation}} \\
\midrule
\textbf{RECAST} & \textbf{9.51} & \textbf{6.20} & \textbf{3.30} & \textbf{0.896} & \textbf{0.909} \\
w/o Sim & 10.39 & 6.76 & 3.64 & 0.883 & 0.896 \\
w/o Unc & 10.36 & 6.78 & 3.58 & 0.869 & 0.888 \\
Only Recent & 10.40 & 6.76 & 3.64 & 0.872 & 0.887 \\
\bottomrule
\end{tabular}
\end{table}

\label{sec:ablation-hparams}
Table~\ref{tab:ablation-pivot} examines how sensitive RECAST is to key hyperparameters on PulseDB.
We use a one-factor-at-a-time sweep: when we vary one parameter, the others are fixed to the best-performing configuration ($\lambda_t=0.1$, $\tau_{\mathrm{sim}}=0.8$, $\tau_{\mathrm{unc}}=0.9$).
For each setting, we report MAE and Pearson correlation for SBP and DBP, and bold indicates the best result.

\textit{Effect of temporal decay ($\lambda_t$):}
Table~\ref{tab:ablation-pivot} (A) shows that recency weighting matters for personalization.
When we \emph{decrease} $\lambda_t$ (weaker decay), Total MAE increases by +0.49 (+5.15\%).
With a further decrease, the degradation becomes larger (+0.99, +10.41\%), and correlations also drop.
This pattern is consistent with the drift setting in streaming biosignals: if $\lambda_t$ is too small, older samples keep large weights, and the update can be influenced by outdated contexts.
In our sweep, $\lambda_t=0.1$ gives a good balance.
It prioritizes recent samples while still allowing the model to use buffered history for stable adaptation.

\textit{Effect of similarity threshold ($\tau_{\mathrm{sim}}$):}
Table~\ref{tab:ablation-pivot} (B) shows that the similarity threshold affects how well RECAST avoids recent but context-mismatched samples.
When we lower $\tau_{\mathrm{sim}}$, Total MAE increases (+0.83, +8.73\%) and correlation decreases, indicating that recency alone is not sufficient for adaptation.
On the other hand, an overly strict threshold ($\tau_{\mathrm{sim}}=0.9$) slightly worsens performance, likely because too few candidates remain to form a representative batch.
In our sweep, $\tau_{\mathrm{sim}}=0.8$ provides the best balance between contextual alignment and sample availability.

\textit{Effect of uncertainty threshold ($\tau_{\mathrm{unc}}$):}
Table~\ref{tab:ablation-pivot} (C) indicates that uncertainty filtering improves reliability, but the sensitivity is smaller than that of decay and similarity.
As $\tau_{\mathrm{unc}}$ decreases (stricter filtering), Total MAE increases (+0.24, +2.52\%) with modest correlation drops.
This suggests a simple quality--quantity trade-off: stricter thresholds remove unreliable candidates, but they can also reduce the pool too much, especially under sparse supervision.
A relatively permissive setting ($\tau_{\mathrm{unc}}=0.9$) works best in our setup, providing reliability control without shrinking the candidate pool excessively.

\textit{Component ablation:}
Table~\ref{tab:ablation-pivot} (D) ablates RECAST's sample selection.
\textsf{w/o Sim} drops the similarity criterion, selecting by $\lambda_t=0.1$ and $\tau_{\mathrm{unc}}=0.9$.
\textsf{w/o Unc} drops the uncertainty filter, selecting by $\lambda_t=0.1$ and $\tau_{\mathrm{sim}}=0.8$.
Both increase Total MAE relative to the full model ($9.51 \to 10.39$ and $10.36$, respectively).
\textsf{Only Recent} goes further, forming each step from only the single most recent labeled and unlabeled sample, and is worse still ($10.40$).
This confirms that recency alone is insufficient and that the scoring criteria contribute complementary gains.
The MC-MED counterpart of this sweep is reported on Table~\ref{tab:ablation-pivot-mcmed} in Appendix~\ref{app:mcmed-sensitivity}.

\subsubsection{Computational Cost and Practicality.}
We measure the per-segment runtime overhead of test-time adaptation on a single GPU with one CPU core.
Table~\ref{tab:runtime-cost} reports, for each method, the latency in milliseconds and as a ratio to the segment duration.
Overall, RECAST costs more per segment than the baselines, but its latency stays well within the data-acquisition window.

\begin{table}[h]
\centering
\caption{Runtime cost per segment on a single GPU and one CPU core. \emph{Ratio (\%)} is the per-segment latency divided by the segment duration.
}
\Description{Per-segment runtime in milliseconds and as a percentage of the segment duration, for No TTA, TTC, and RECAST on PulseDB (10\,s segments) and MC-MED (60\,s segments). RECAST adds modest latency, around 4\% of the window on PulseDB and under 1\% on MC-MED.}
\label{tab:runtime-cost}
\begin{tabular}{l|rc|rc}
\toprule
\multirow{2}{*}{\textbf{Method}} & \multicolumn{2}{c|}{\textbf{PulseDB} (10\,s)} & \multicolumn{2}{c}{\textbf{MC-MED} (60\,s)} \\
 & \multicolumn{1}{c}{\textbf{ms}} & \textbf{Ratio (\%)} & \multicolumn{1}{c}{\textbf{ms}} & \textbf{Ratio (\%)} \\
\midrule
No TTA & 40.35 & 0.40 & 79.78 & 0.13 \\
TTC & 266.27 & 2.66 & 314.41 & 0.52 \\
RECAST & 428.87 & 4.29 & 435.79 & 0.73 \\
\bottomrule
\end{tabular}
\end{table}

In detail, online adaptation methods increase latency over No TTA on both datasets, and MC-MED has a higher inference-only cost because its 60\,s segments require more computation per forward pass.
RECAST's total runtime is nonetheless similar on PulseDB and MC-MED ($428.9$ vs.\ $435.8$\,ms), indicating that latency is dominated by the adaptation procedure (buffer processing, scoring, and update) rather than the forward pass on the raw segment.
Its per-sample scoring (latent-space similarity and predictive uncertainty) raises latency above TTC by $61\%$ on PulseDB ($429$ vs.\ $266$\,ms) and $39\%$ on MC-MED ($436$ vs.\ $314$\,ms).
In all cases, however, the per-segment latency stays well below 1\,s, consuming about 4\% of the segment window on PulseDB and 0.7\% on MC-MED.

The dominant time cost in any deployment is offline pre-training (orders of magnitude longer than per-step adaptation) and physical signal acquisition, which paces the pipeline.
Against these scales, RECAST's added per-step cost is negligible, and its accuracy gains come at low computational cost.

\begin{figure*}[t]
\centering
\begin{subfigure}{0.48\linewidth}
\centering
\includegraphics[width=\linewidth]{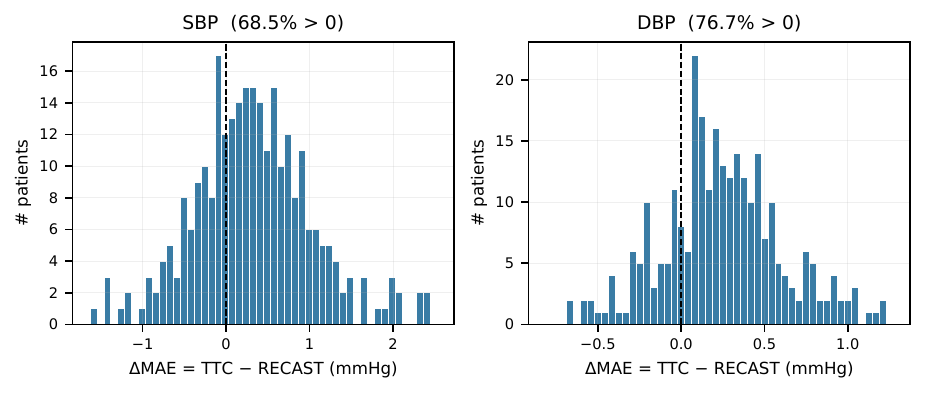}
\caption{PulseDB ($n=279$ sessions)}
\label{fig:app-hist-pulsedb}
\end{subfigure}
\hfill
\begin{subfigure}{0.48\linewidth}
\centering
\includegraphics[width=\linewidth]{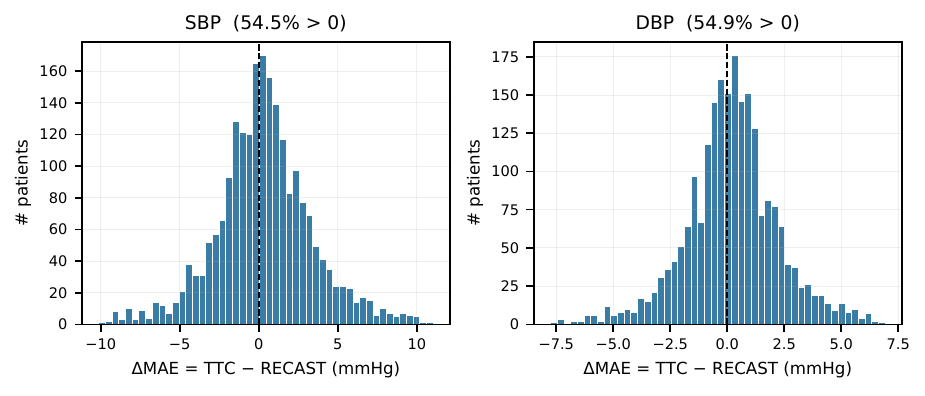}
\caption{MC-MED ($n=2{,}244$ patients)}
\label{fig:app-hist-mcmed}
\end{subfigure}
\caption{Per-patient $\Delta$MAE (MAE\textsubscript{TTC} $-$ MAE\textsubscript{RECAST}) distribution, with separate SBP and DBP panels. Positive values indicate RECAST improvement.}
\Description{Per-patient delta-MAE histograms for PulseDB (left) and MC-MED (right), each with an SBP and a DBP panel. Most patients lie on the improvement side, more so on PulseDB.}
\label{fig:app-hist}
\end{figure*}

\begin{figure*}[t]
\centering
\begin{subfigure}{0.48\linewidth}
\centering
\includegraphics[width=\linewidth]{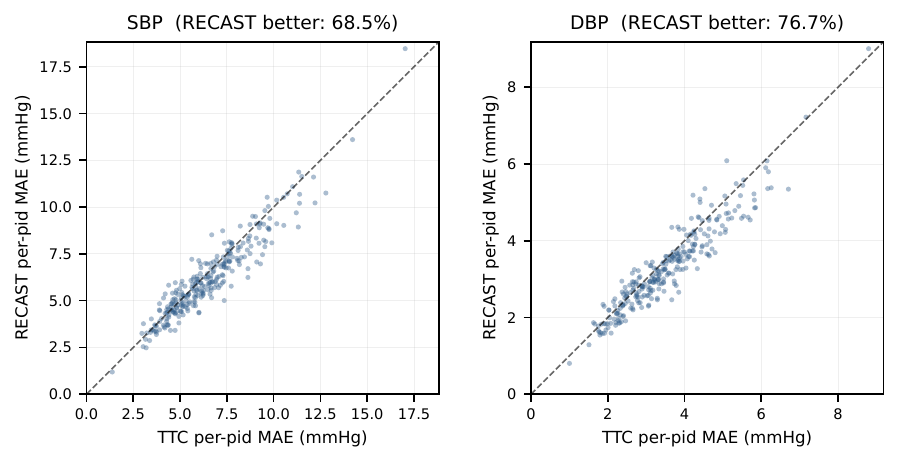}
\caption{PulseDB ($n=279$ sessions)}
\label{fig:app-scatter-pulsedb}
\end{subfigure}
\hfill
\begin{subfigure}{0.48\linewidth}
\centering
\includegraphics[width=\linewidth]{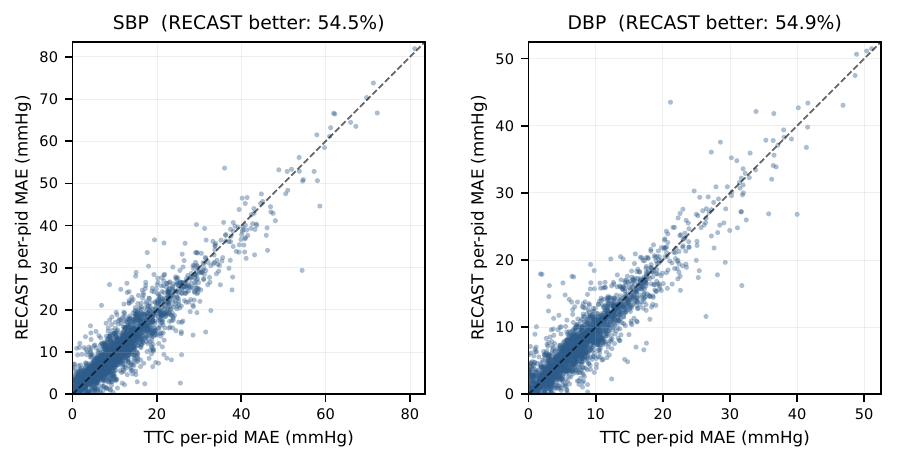}
\caption{MC-MED ($n=2{,}244$ patients)}
\label{fig:app-scatter-mcmed}
\end{subfigure}
\caption{Per-patient MAE scatter: TTC on the $x$-axis vs.\ RECAST on the $y$-axis, with separate SBP and DBP panels. Points below the diagonal indicate RECAST improvement.}
\Description{Per-patient MAE scatter for PulseDB (left) and MC-MED (right), each with an SBP and a DBP panel. Most points lie below the diagonal and the gap widens with TTC error.}
\label{fig:app-scatter}
\end{figure*}

\section{Conclusion}
\label{sec:conclusion}
We study test-time personalization for streaming biosignal-based BP estimation, and we focus on the role of \emph{sample selection} for stable adaptation.
Streaming biosignals are non-stationary and can also show abrupt regime changes.
In this situation, simple buffer usage strategies, such as using all history uniformly or updating from only the most recent sample, can be suboptimal.

We propose \textbf{RECAST} (REcent \& Context-Aware Sampling for TTA), a lightweight plug-in sampling module for buffered TTA frameworks.
RECAST constructs adaptation batches using temporal recency, contextual similarity, and predictive reliability, requiring no architectural changes to the host framework.
On PulseDB and MC-MED, RECAST is the most accurate of the compared methods, attaining the lowest MAE and RMSE on all three targets, including on the heterogeneous emergency-department stream.
Per-patient analysis shows that the benefit is uneven: on PulseDB the majority of sessions improve, whereas on MC-MED the gains concentrate on the hardest and most variable patients, since most MC-MED patients vary little within a stay. Improvements are statistically significant on both datasets.
Ablations indicate the three criteria are complementary: recency alone is insufficient, and contextual similarity carries most of the gain, most clearly on MC-MED.
At the same time, RECAST remains practical, with sub-second per-segment latency on a single GPU with one CPU core, well below the natural data-acquisition window.
More broadly, our findings suggest that sample selection is a decisive yet under-specified knob in buffered streaming TTA.
Making it explicit yields consistent gains without changing the model or its training objective.

Our study has limitations.
RECAST measures recency by buffer insertion order, which may not match real elapsed time under irregular supervision, and its benefit depends on within-patient variation, offering little for near-stationary patients (Section~\ref{sec:variability}).
We also instantiate it only within the TTC framework, though it applies in principle to other buffered TTA settings.


\appendix

\begin{table*}[t]
\centering
\caption{Per-target within-patient-variability breakdown (SBP and DBP), pooled across PulseDB and MC-MED. Quintiles are by within-patient SBP std (Q1 most stable to Q5 most variable). The \emph{within-pt std} columns are the median per-patient standard deviation of reference SBP and DBP, in mmHg. \emph{\% imp.}\ is the fraction of patients RECAST improves over TTC, and \emph{gain} is the median relative improvement, for SBP and DBP. The bottom rows reconcile the quintiled counts with the full evaluation set.}
\Description{Per-quintile table of RECAST's SBP and DBP improvement against within-patient SBP variability. SBP and DBP improve fewest patients in the most stable quintile and rise to 55 to 63 percent improved with about +4 percent median gain in the higher-variability quintiles. The bottom rows show 852 single-reading MC-MED patients are excluded, so 1,392 quintiled plus 852 equal the 2,244 evaluated patients.}
\label{tab:variability-target}
\begin{tabular}{c|cc|cc|cc|cc}
\toprule
\multirow{2}{*}{\textbf{Quintile}} & \multicolumn{2}{c|}{\textbf{within-pt std}} & \multicolumn{2}{c|}{\textbf{SBP}} & \multicolumn{2}{c|}{\textbf{DBP}} & \multicolumn{2}{c}{\textbf{\# patients}} \\
\cmidrule(lr){2-3}\cmidrule(lr){4-5}\cmidrule(lr){6-7}\cmidrule(lr){8-9}
 & \textbf{SBP} & \textbf{DBP} & \textbf{\% imp.} & \textbf{gain (\%)} & \textbf{\% imp.} & \textbf{gain (\%)} & MC-MED & PulseDB \\
\midrule
Excluded & --- & --- & --- & --- & --- & --- & 852 & 0 \\
Q1 (most stable) & 2.7 & 4.2 & 51.6 & $+1.3$ & 51.9 & $+1.2$ & 336 & 1 \\
Q2 & 6.2 & 5.0 & 54.5 & $+3.1$ & 59.6 & $+4.2$ & 310 & 22 \\
Q3 & 9.2 & 5.9 & 61.1 & $+4.3$ & 62.9 & $+5.6$ & 266 & 68 \\
Q4 & 12.5 & 7.7 & 59.9 & $+3.9$ & 61.7 & $+5.1$ & 244 & 90 \\
Q5 (most variable) & 18.6 & 9.0 & 60.2 & $+4.2$ & 60.2 & $+3.6$ & 236 & 98 \\
\midrule
Total & & & & &  & & 2{,}244 & 279 \\
\bottomrule
\end{tabular}
\end{table*}

\section{Appendix: Per-Target Per-Patient Breakdown}
\label{app:per-patient}

This appendix breaks down the main-text per-patient analysis by target, reporting SBP and DBP separately.
It confirms that the pattern reported for Total in the main paper (Figures~\ref{fig:per-patient} and~\ref{fig:pp-scatter}) holds for each BP component, not only in aggregate.

Figure~\ref{fig:app-hist} is the per-patient $\Delta$MAE distribution, with separate SBP and DBP panels; bars to the right of zero are patients RECAST improves.
RECAST improves the majority on both targets: SBP error for $68.5\%$ (PulseDB) and $54.5\%$ (MC-MED) of patients, and DBP error for $76.7\%$ and $54.9\%$.
On PulseDB the bars sit mostly to the right of zero, with DBP improving slightly more sessions than SBP; on MC-MED both are more balanced but remain majority-positive.

Figure~\ref{fig:app-scatter} is the per-patient MAE scatter, TTC on the $x$-axis against RECAST on the $y$-axis, again with separate SBP and DBP panels.
For both targets, points lie mostly below the diagonal, and the gap widens as the baseline TTC error grows, the same difficulty dependence seen for Total.
RECAST's behavior is therefore not driven by a single BP component.

\section{Appendix: Within-Patient Variability, by Target}
\label{app:variability}

This appendix breaks down the within-patient-variability analysis of Section~\ref{sec:variability} by target, reporting SBP and DBP separately.
Computing a within-patient standard deviation requires at least two reference labels.
Table~\ref{tab:variability-target} reports, per quintile, the median within-patient SBP and DBP std and RECAST's improvement on SBP and DBP separately.

SBP and DBP individually follow the Total pattern from Section~\ref{sec:variability}.
In the most stable quintile, improvement is smallest for both (SBP $51.6\%$/$+1.3\%$, DBP $51.9\%$/$+1.2\%$).
From the second quintile on, both rise to about $55$--$63\%$ improved with $+3$ to $+6\%$ median gain.
The effect is therefore not specific to either target.
The relationship is a threshold rather than a monotonic trend: a modest amount of within-patient variation is enough to unlock the gain, after which it saturates, since very high variability also includes abrupt transitions that no buffered method fully tracks.

\begin{table}[t]
\centering
\caption{Ablation study on RECAST components and key hyperparameters on MC-MED. Best results in each block are boldfaced.}
\Description{Ablation table on MC-MED with four blocks: (A) temporal decay rate, (B) similarity threshold, (C) uncertainty threshold, and (D) component ablation. Each block reports total, SBP, and DBP MAE and SBP and DBP correlation. The chosen configuration is best in the decay and similarity sweeps; the uncertainty threshold has negligible effect, while removing similarity or using only the most recent sample degrades performance.}
\label{tab:ablation-pivot-mcmed}
\begin{tabular}{c|ccc|cc}
\toprule
\textbf{Setting} & \multicolumn{3}{c|}{\textbf{MAE $\downarrow$}} & \multicolumn{2}{c}{\textbf{Correlation $\uparrow$}} \\
& \textbf{Total} & \textbf{SBP} & \textbf{DBP} & \textbf{SBP} & \textbf{DBP} \\
\midrule
\multicolumn{6}{l}{\textbf{(A) Temporal decay rate} $\boldsymbol{\lambda_t}$} \\
\midrule
\textbf{0.10} & \textbf{19.44} & \textbf{11.22} & \textbf{8.22} & \textbf{0.791} & \textbf{0.699} \\
0.05 & 19.92 & 11.51 & 8.41 & 0.777 & 0.689 \\
0.01 & 20.58 & 11.98 & 8.60 & 0.762 & 0.685 \\
\midrule
\multicolumn{6}{l}{\textbf{(B) Similarity threshold} $\boldsymbol{\tau_{\mathrm{sim}}}$} \\
\midrule
\textbf{0.9} & \textbf{19.44} & \textbf{11.22} & \textbf{8.22} & \textbf{0.791} & \textbf{0.699} \\
0.8 & 19.89 & 11.49 & 8.40 & 0.780 & 0.693 \\
0.7 & 20.10 & 11.65 & 8.45 & 0.776 & 0.693 \\
0.6 & 20.21 & 11.72 & 8.49 & 0.774 & 0.689 \\
0.5 & 20.32 & 11.81 & 8.51 & 0.772 & 0.693 \\
\midrule
\multicolumn{6}{l}{\textbf{(C) Uncertainty threshold} $\boldsymbol{\tau_{\mathrm{unc}}}$ \textnormal{\footnotesize(no effect at $\tau_{\mathrm{sim}}{=}0.9$)}} \\
\midrule
\textbf{0.9} & \textbf{19.44} & \textbf{11.22} & \textbf{8.22} & \textbf{0.791} & \textbf{0.699} \\
0.8 & 19.44 & 11.22 & 8.22 & 0.791 & 0.699 \\
0.7 & 19.44 & 11.22 & 8.22 & 0.791 & 0.699 \\
0.6 & 19.44 & 11.22 & 8.22 & 0.791 & 0.699 \\
0.5 & 19.44 & 11.22 & 8.22 & 0.791 & 0.699 \\
\midrule
\multicolumn{6}{l}{\textbf{(D) Component ablation}} \\
\midrule
RECAST & 19.44 & 11.22 & 8.22 & 0.791 & 0.699 \\
w/o Sim & 20.28 & 11.79 & 8.49 & 0.772 & 0.694 \\
\textbf{w/o Unc} & \textbf{19.43} & \textbf{11.22} & \textbf{8.21} & \textbf{0.791} & \textbf{0.699} \\
Only Recent & 19.54 & 11.28 & 8.26 & 0.790 & 0.695 \\
\bottomrule
\end{tabular}
\end{table}
\section{Appendix: MC-MED Hyperparameter Sensitivity}
\label{app:mcmed-sensitivity}

This appendix extends the hyperparameter-sensitivity analysis to MC-MED.
Following the same one-factor-at-a-time structure as the PulseDB sweep (Table~\ref{tab:ablation-pivot}), each block varies one parameter while holding the other two at the best MC-MED configuration ($\tau_{\mathrm{sim}}{=}0.9$, $\lambda_t{=}0.1$, $\tau_{\mathrm{unc}}{=}0.9$).
Table~\ref{tab:ablation-pivot-mcmed} reports, for each block, the total, SBP, and DBP MAE and the SBP and DBP correlation.

The temporal-decay and similarity sweeps follow the expected direction.
In block (A), weakening the decay (more weight on older samples) raises Total MAE monotonically, from $19.44$ at $\lambda_t{=}0.1$ to $20.58$ at $\lambda_t{=}0.01$, confirming that recency matters.
In block (B), loosening the similarity threshold steadily degrades accuracy, from $19.44$ at $\tau_{\mathrm{sim}}{=}0.9$ to $20.32$ at $\tau_{\mathrm{sim}}{=}0.5$, so tighter context matching helps.
The chosen configuration ($\tau_{\mathrm{sim}}{=}0.9$, $\lambda_t{=}0.1$, $\tau_{\mathrm{unc}}{=}0.9$) attains the lowest Total MAE ($19.44$), matching RECAST in Table~\ref{tab:main-results-expanded}.

Block (C) behaves differently: the rows are identical across $\tau_{\mathrm{unc}}\in[0.5,0.9]$.
This is expected here, not an artifact.
Once the similarity gate is tight ($\tau_{\mathrm{sim}}{=}0.9$), it already removes most off-context candidates, so the uncertainty threshold rarely binds.
Contextual similarity, rather than uncertainty filtering, therefore drives the MC-MED gains, and RECAST is robust to the choice of $\tau_{\mathrm{unc}}$ here.

Block (D) makes this explicit, where \textsf{w/o Sim} and \textsf{w/o Unc} denote disabling the corresponding threshold ($\tau\to 0$).
Removing the similarity filter raises Total MAE from $19.44$ to $20.28$.
Removing the uncertainty filter, in contrast, leaves the error statistically unchanged ($19.43$ vs.\ $19.44$, a $0.01$\,mmHg difference within rounding).
We still set $\tau_{\mathrm{unc}}{=}0.9$ to keep a single RECAST configuration across both datasets, since on PulseDB the same filter does reduce error (Table~\ref{tab:ablation-pivot}).
On MC-MED, similarity gating thus accounts for essentially all of RECAST's benefit, while the uncertainty filter is redundant once the similarity gate is tight.
The recency-only variant \textsf{Only Recent}, which adapts on only the single most recent labeled and unlabeled sample, reaches $19.54$, slightly behind the full model.

\newpage
\bibliographystyle{ACM-Reference-Format}
\bibliography{sample-base}

\end{document}